%% file: main.tex
\documentclass[letterpaper, 10 pt, conference]{ieeeconf}  %

\IEEEoverridecommandlockouts                              %

\usepackage[backend=biber,
            hyperref=true,
            url=false,
            isbn=false,
            doi=false,
            backref=false,
            style=ieee,
            citestyle=numeric-comp,
            sorting=none,%
            block=none]{biblatex}

\usepackage{xcolor}
\usepackage{xspace}
\usepackage{graphicx}
\usepackage{booktabs}
\usepackage{lipsum}
\usepackage{url}
\usepackage{svg}
\usepackage{hyperref}
\usepackage{multirow}
\usepackage{adjustbox}
\usepackage{xcolor} %
\usepackage{listings}
\definecolor{lightgray}{rgb}{0.95, 0.95, 0.95}
\definecolor{highlight}{rgb}{0.0, 0.0, 0.81} %

\hypersetup{
    colorlinks=false,    %
    linkcolor=blue,      %
    citecolor=blue,      %
    filecolor=magenta,   %
    urlcolor=cyan,       %
    pdfborder={0 0 0.2}, %
}

\usepackage{capt-of}

\newcommand{\acro}{BUMBLE}

\newcommand{\mycomment}[1]{}

\title{\LARGE \bf BUMBLE: Unifying Reasoning and Acting with\\Vision-Language Models for Building-wide Mobile Manipulation}
\author{Rutav Shah\quad Albert Yu\quad Yifeng Zhu\quad Yuke Zhu$^{*}$\quad Roberto Mart{\'i}n-Mart{\'i}n$^{*}$%
\\
The University of Texas at Austin
\thanks{$^{*}$ Equal advising. Correspondence: rutavms@utexas.edu}%
}

\begin{document}

\maketitle
\thispagestyle{empty}
\pagestyle{empty}

\begin{abstract}
To operate at a building scale, service robots must perform very long-horizon mobile manipulation tasks by navigating to different rooms, accessing different floors, and interacting with a wide and unseen range of everyday objects.
We refer to these tasks as Building-wide Mobile Manipulation.
To tackle these inherently long-horizon tasks, we introduce~\acro{}, a unified Vision-Language Model (VLM)-based framework integrating open-world RGBD perception, a wide spectrum of gross-to-fine motor skills, and dual-layered memory.
Our extensive evaluation ($90+$ hours) indicates that \acro{} outperforms multiple baselines in long-horizon building-wide tasks that require sequencing up to $12$ ground truth skills spanning $15$ minutes per trial. 
\acro{} achieves $47.1\%$ success rate averaged over $70$ trials in different buildings, tasks, and scene layouts from different starting rooms and floors. Our user study demonstrates $22\%$ higher satisfaction with our method than state-of-the-art mobile manipulation methods.
Finally, we demonstrate the potential of using increasingly-capable foundation models to push performance further. For more information, see \textcolor{blue}{\href{https://robin-lab.cs.utexas.edu/BUMBLE/}{https://robin-lab.cs.utexas.edu/BUMBLE/}}
\end{abstract}

\input{00_introductionv3}

\input{01_related_works}

\input{02_method}

\input{04_results}

\input{05_conclusion}

\printbibliography
\input{appendix}
\end{document}

%% file: 00_introductionv3.tex
\section{Introduction}
\label{s:intro}

\begin{figure}[t!]
    \centering
    \includegraphics[width=\columnwidth]{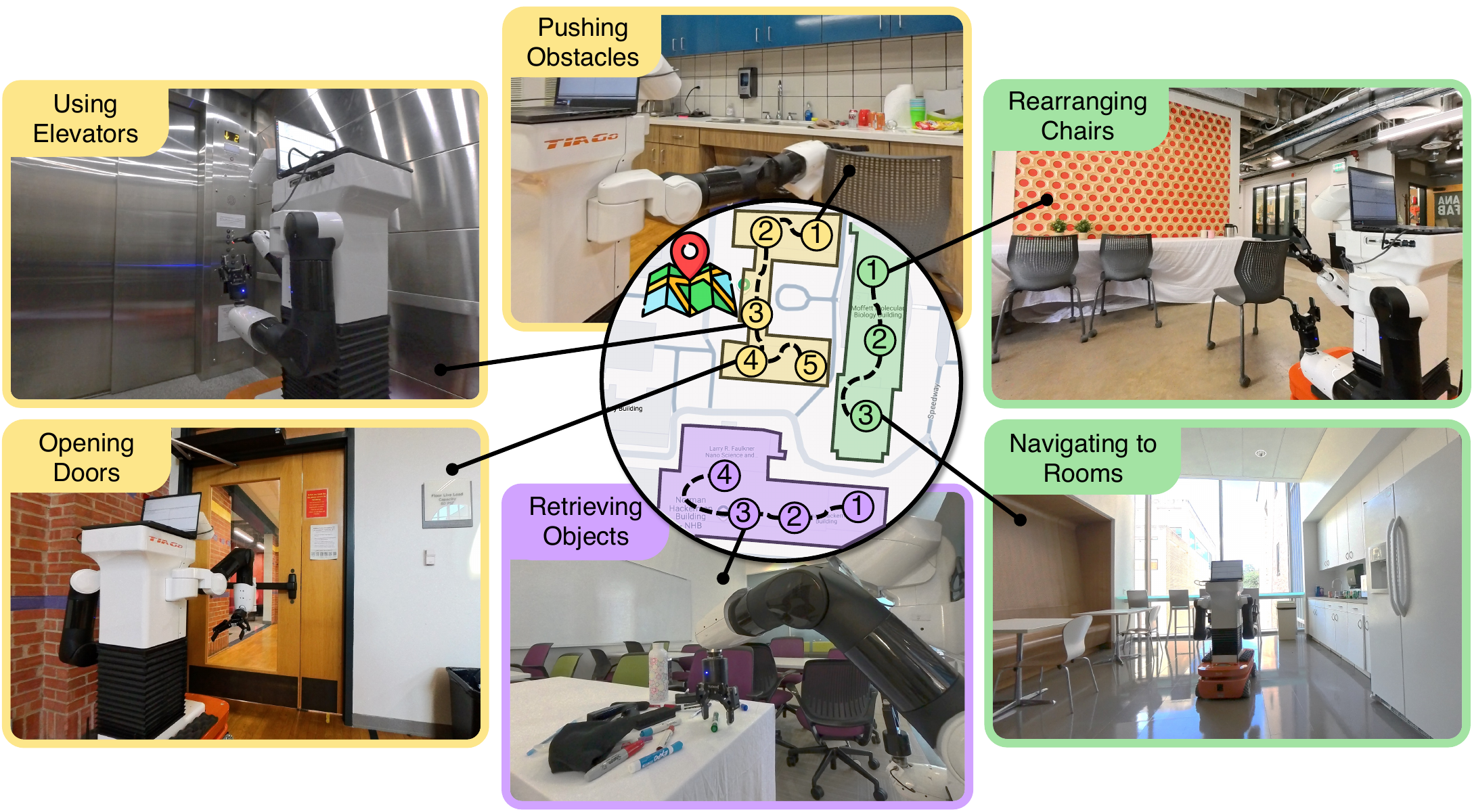}
    \caption{\textbf{Building-wide mobile manipulation}. At a building-wide scale, mobile manipulation tasks involve sequencing multiple skills like pushing obstacles or opening doors to clear robot pathways, using elevators to reach a destination floor, rearranging chairs in the workspace, and retrieving objects. We present a framework, \acro{}, that can solve long-horizon, spatially expansive tasks across different buildings.}
    \label{fig:pull_figure}
\end{figure}

Autonomous service robots aiming to assist humans daily must perform mobile manipulation (MoMa) in homes, hospitals, universities, offices, and other building-wide environments.
Given a user instruction, like ``I am on a diet, but I want soda,'' the robot must interpret the free-form instruction, create a high-level task plan to accomplish the task, and instantiate low-level robot commands to fulfill it.
To accomplish such tasks in buildings, the robot might have to traverse to a kitchen, potentially using an elevator if it is on a different floor.
To use the elevator, the robot must draw on prior experiences with elevators in other buildings.
The robot must also handle unexpected obstacles, such as by pushing items blocking a narrow corridor or opening doors to move in and out of the room.
Finally, even after reaching the kitchen, the robot must identify an appropriate soda can from the diverse clutter often present in human-inhabited buildings, recognizing the ``diet'' option even if it has never been encountered before.
We refer to such long-horizon and spatially expansive tasks as \textbf{Building-wide Mobile Manipulation} (see Fig.~\ref{fig:pull_figure}) and present a unified framework to address it.

\begin{figure*}[t]
    \centering
    \includegraphics[width=\textwidth]{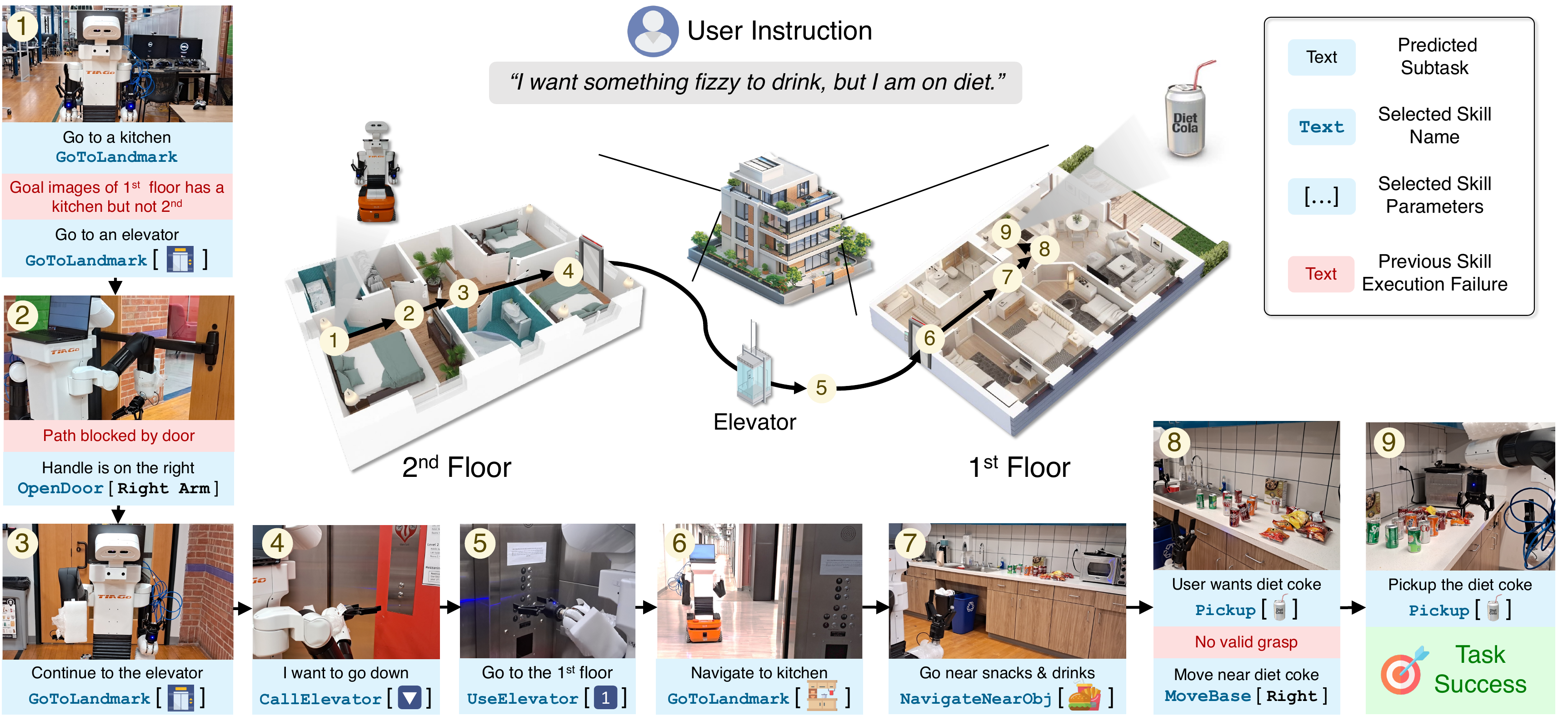}
    \vspace{-10px}
    \caption{\textbf{Building-wide Mobile Manipulation} tasks require navigating various rooms and floors while interacting with diverse objects. To solve such long-horizon tasks, \acro{} leverages a diverse skill library enabling navigation and interaction, with parameterized skills adaptable to different scenes. At each skill execution step (circled numbers), \acro{} uses VLM's reasoning capabilities to select the next skill (\textcolor[HTML]{00628F}{blue text}) and skill parameters (\textbf{[\ldots]}) and recover from failures (\textcolor[HTML]{AF0002}{red text}) to solve building-wide tasks effectively.}
    \label{fig:figure2}
\end{figure*}
Building-wide MoMa entails a series of technical challenges, including integrating complex locomotion and manipulation behaviors in buildings, perceiving objects and their context in the open-world, and using past experiences to learn from mistakes reason about long-horizon strategies (example execution in Fig.~\ref{fig:figure2}).
Decades of research in Task and Motion Planning (TAMP)~\cite{kaelbling2011hierarchical,srivastava2014combined,garrett2021integrated,wolfe2010combined} and recent neuro-symbolic planning~\cite{xu2018neuraltaskprogramminglearning,xu2019regressionplanningnetworks,mao2023pdsketch,9561548} have been applied to long-horizon robotic tasks. However, these approaches require pre-defined symbolic abstractions and domain knowledge, confining their reasoning capabilities to a closed set of objects.
Large Language Models (LLMs)~\cite{brohan2023can, huang2022language, brohan2023can, liang2023code, rana2023sayplan, hu2024deploying, honerkamp2024language} are capable of reasoning about open-world scenarios to create task plans. However, they lack grounded and direct reasoning through visual perception, inherently falling short in the precise geometric reasoning crucial for producing executable robot actions.
Recently, Vision-Language Models (VLMs) have emerged as a promising tool for Building-wide MoMa, unlocking grounded reasoning in open-world scenes~\cite{hu2023look,liu2024moka,huang2024copa,chang2023data,chaplot2020object,majumdar2022zson,huang2023visual,jagan2024convoi,liu2024dragon}. 
Some works~\cite{homerobot,liu2024okrobot,stone2023open,qiu2024open,zhi2024closed} have demonstrated the potential of VLMs in simple MoMa tasks but lack the necessary memory and motor skills to operate at a building scale. In contrast, building-wide MoMa requires 1) an open-world perception system for reasoning about diverse objects, 2) complex motor skills to act effectively in buildings, and 3) memory for temporally extended reasoning in long-horizon task execution.

We propose to tackle Building-wide MoMa by unifying reasoning and acting through a VLM-based framework---one that can handle diverse objects with open-world perception, act effectively in building-wide situations with a broad set of gross-to-fine motor skills, and learn and adapt from past experience through memory.
By seamlessly unifying these capabilities into rapidly advancing VLMs, our general framework improves as VLM backbones become better at sophisticated perception and reasoning.

To this end, we introduce \acro{} (\textbf{BU}ilding-wide \textbf{M}o\textbf{B}i\textbf{LE} Manipulation), a new framework that addresses all of these needs with an adaptable approach capable of reasoning and acting over new objects, tasks, and buildings.
Given a task instruction in free-form language, \acro{} reasons over the scene, past experiences, and motor skill capabilities to predict the next parameterized skill to execute. \acro{} comprises four key ingredients: a) a \textbf{VLM} serving as the central reasoning module connecting perception, memory, and skills; b) \textbf{dual-layered memory}: short-term memory~\cite{cowan2014working} to maintain robot \textit{execution history}, and long-term memory~\cite{shiffrin1968search} to store valuable experience and concepts from \textit{past trials}; c) a diverse \textbf{skill library} of parameterized skills that enable the robot to navigate to different floors and rooms (e.g., \texttt{GoToLandmark}[\texttt{GoalImage}], \texttt{UseElevator}[\texttt{Button}]), adjust the robot base for manipulation (e.g., \texttt{MoveBase}[\texttt{Dir}.]), and interact with objects through diverse contact-rich behaviors (e.g., \texttt{PushObjOnGround}[\texttt{ObjSeg}., \texttt{Dir}.]) present in different buildings; d) \textbf{perception system} for the VLM to not only visually reason about open-world, cluttered scenes but also process depth information necessary for embodied decisions.

Our experiments demonstrate that \acro{} can successfully execute long-horizon, building-wide tasks requiring up to $12$ parameterized skills and over $15$ minutes per trial (excluding VLM query time). These tasks involve navigating different floors and rooms, clearing pathway obstacles, retrieving objects, and making granular decisions, such as adjusting the base for manipulation.
Notably, \acro{} achieves a $47.1\%$ average success rate in tasks like retrieving soda cans and rearranging chairs across three buildings.
We demonstrate that \acro{} solves previously unseen tasks involving multiple plausible solutions and operating in completely new rooms. Through a user study, we find that \acro{} rollouts align with human preferences $22\%$ better than prior state-of-the-art VLM-based MoMa approaches.
Furthermore, we demonstrate how \acro{} improves performance with advancing VLM capabilities.
Finally, we thoroughly analyze failures, guiding future research.

%% file: 01_related_works.tex
\section{Related Work}
\label{s:rw}
\begin{figure*}
    \centering
    \includegraphics[width=\linewidth]{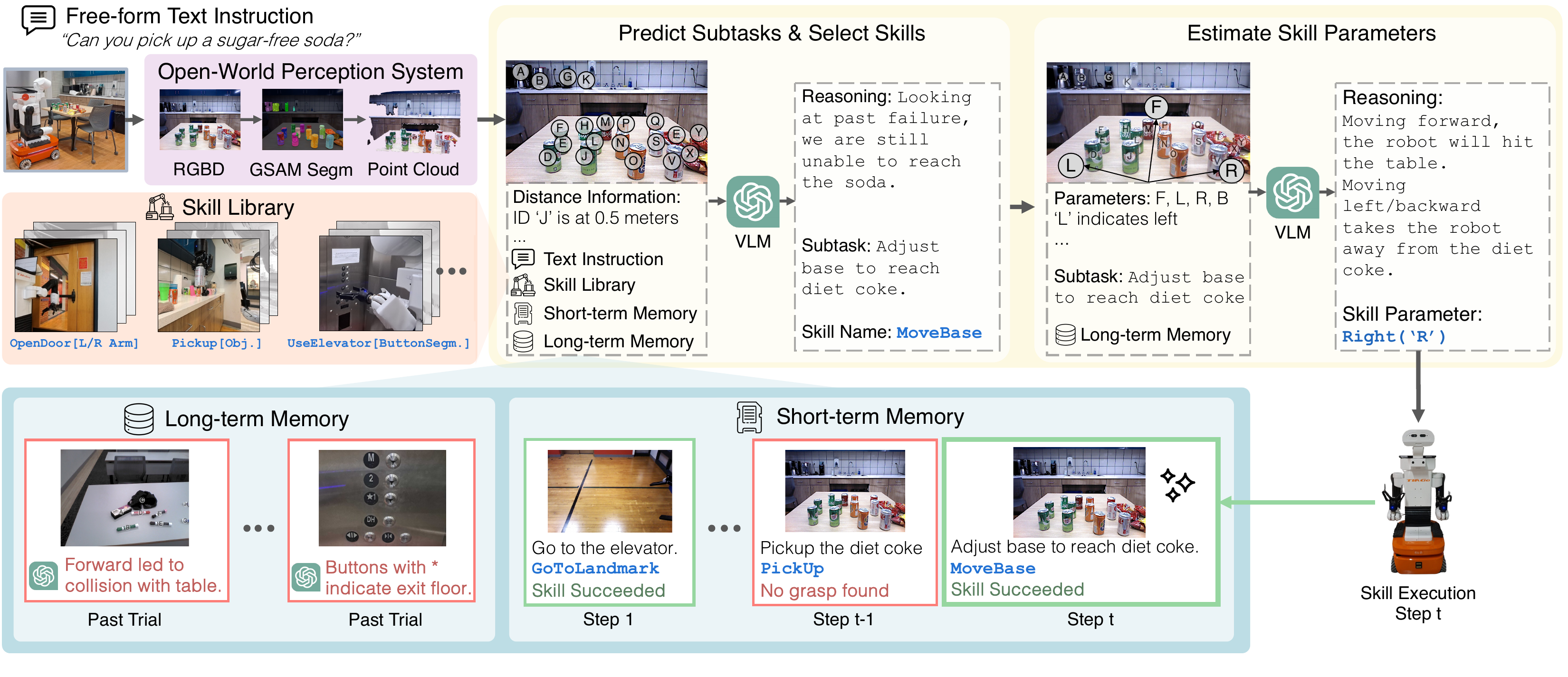}
    \vspace{-20px}
    \caption{
    \textbf{BUMBLE Architecture}. Given a free-form text instruction, skill library (\textit{top Left}), and short- and long-term memory (\textit{bottom middle and left}), ~\acro{} iteratively perceives the environment through onboard RBGD sensors (\textit{top left}), predicts parameterized skills (\textit{top middle and right}), and executes them in the environment.
    The predicted skill and its parameters are executed and stored in short-term memory for iterative prediction (\textit{bottom right}).
    }
    \label{fig:method}
\end{figure*}
Pioneering research on \textbf{mobile manipulators}~\cite{schuler1987integration,dario1993urmad,khatib1996coordination,ambrose2004mobile,katz2006umass} performed tasks in human environments, which naturally occur in buildings~\cite{khatib1999mobile,bohren2011towards,ciocarlie2012mobile}. 
Early work in whole-body control~\cite{tanner2003nonholonomic,brock2002task,peterson2000high,seraji1998unified,holmberg2000development} and motion planning~\cite{huang2000coordinated,nagatani2002motion,oriolo2005motion,li2020hierarchical,sandakalum2022motion} laid the groundwork for low-level control in building-level mobile manipulation, but did not address reasoning required in long-horizon building-wide tasks.
Task and motion planning (TAMP)~\cite{kaelbling2011hierarchical,srivastava2014combined,garrett2021integrated,wolfe2010combined,bajracharya2024demonstrating} integrates reasoning for task and motion generation to achieve long-horizon tasks. However, it is confined to a fixed preprogrammed set of object categories and scenes, hindering its application to building-scale environments with open-world objects.

Recently, \textbf{foundation models}~\cite{bommasani2021opportunities} have been leveraged for mobile manipulation. LLMs excel at semantic reasoning for planning~\cite{ding2023integrating,lin2023text2motion}, but their predictions are not grounded in the scene, requiring additional machinery to match their plans to executable robot actions~\cite{singh2023progprompt,liang2023code,brohan2023can,wang2024llm}.
VLMs provide a more promising engine to unify and ground reasoning for tabletop manipulation~\cite{stone2023open,hu2023look,liu2024moka,huang2024copa},  navigation~\cite{chaplot2020object,majumdar2022zson,huang2023visual,jagan2024convoi,nasiriany2024pivot}, and mobile manipulation~\cite{qiu2024open,zhi2024closed,wang2024mosaic,homerobot,liu2024okrobot}, but prior methods lack a diverse skill repertoire (navigation and manipulation, coarse and fine motion) and long-horizon memory-based reasoning capability, both of which are necessary to perform building-wide mobile manipulation tasks. \acro{} is a natural extension of VLM-based robotic systems: it uses VLMs to unify semantic and geometric reasoning for manipulation and navigation at the building-wide scale, enabling it to leverage current and future advances in VLM models. Our framework integrates perceptual and motor skills and long and short-term memory to obtain efficiency and generalization to diverse objects and scenes in different buildings.

All aforementioned prior work in mobile manipulation assumes the navigation path is either obstacle-free or that obstacles can be navigated around.
In \textbf{interactive navigation}~\cite{salomon2003interactive,xia2020interactive} however, obstacles must be moved or manipulated---such as pushing away path-blocking boxes, opening doors, or pressing elevator buttons---to reach a destination.
Prior methods tackled this with motion planning~\cite{stilman2005navigation,van2008interactive,wang2020affordance} and learning~\cite{hrl4in,zeng2021pushing,wang2024camp} but focused solely on the interactive navigation problem using only geometric information.
In contrast, \acro{} integrates both semantic reasoning (e.g., avoiding pushing delicate objects) and geometric reasoning (e.g., avoiding object collisions while pushing an obstacle) into a unified method for mobile manipulation.

%% file: 02_method.tex
\section{\acro{}: VLM-Based Building-Wide MoMa}
\label{sec:method}
\acro{} is a new framework integrating perception, motor skills, and memory in a VLM-based solution for high-level reasoning and low-level control of a mobile manipulator for building-wide tasks.
In this section, we describe the key technical blocks of~\acro{} (see Fig.~\ref{fig:method}): (a) capabilities for open-world perception, a library of diverse skills to act in the physical world, and short and long-term memory for on-the-fly adaptation and learning from past mistakes; and (b) a VLM-based decision-making module to reason and ground the text instruction in the current scene and predict the next parameterized skill to execute (see App.~\ref{app:method} for details).

\subsection{Perception System, Skill Library, and Memory}\label{method:vision_mem}

\paragraph{Open-world Perception System}
Robots must perceive and localize diverse open-world objects to operate effectively in buildings. We address this using a perception system with a robust segmentation model, Grounded-SAM (GSAM)~\cite{kirillov2023segany,liu2023grounding,ren2024grounded}, for segmenting foreground (i.e., interactable) objects (Fig.~\ref{fig:method}, purple).
We use GSAM instead of directly querying the VLM because segmentation models offer pixel-level accuracy, allowing for more precise localization and manipulation. 
After obtaining the object masks for the scene, we calculate the object point cloud by back-projecting depth images and determine the precise distance between the robot and the detected objects.
The object proximity information enables \acro{} to estimate the correct skill for manipulation in building environments, while the object point cloud enables precise manipulation using motor skills.

\paragraph{Skill Library}
To overcome the challenges of building-wide MoMa, \acro{} requires a diverse skill library allowing for high-level abstract behaviors like navigating to a room, down to more fine-grained behaviors like adjusting the base for manipulation (Fig.~\ref{fig:method}, yellow; examples in Fig.~\ref{fig:figure2},~\ref{fig:exec_traces}).
The skill library must also support using building-specific fixtures like elevators and perform manipulation skills like pushing objects and picking up items, addressing interactive navigation challenges~\cite{li2020hierarchical,xia2020interactive}.
To this end, we construct a skill library to which new skills can easily be added and chosen by the VLM.
Our library includes skills such as \texttt{GoToLandmark}, \texttt{NavigateNearObj}, \texttt{MoveBase}, \texttt{Pickup}, \texttt{PushObjOnGround}, \texttt{OpenDoor}, \texttt{CallElevator}, and \texttt{UseElevator}.
The parameters of each skill are based on object or robot configurations, except for \texttt{GoToLandmark}.
It uses a topological visual map~\cite{kovsnar2008visual} of the building, with landmark images as nodes and 2D occupancy maps to generate trajectories between them (See Fig.~\ref{fig:exec_traces}).
With the library of parameterized skills and VLM's reasoning capabilities,~\acro{} can integrate semantic and geometric reasoning at the task level by predicting the subtask and skill and the robot's motion by estimating skill parameters to accomplish building-wide tasks.

\paragraph{Memory} Building-wide MoMa is inherently long-horizon, requiring the robot to track its state-action history.
Moreover, the robot must recover from failed skill executions whenever possible, such as by moving away obstacles or adjusting the robot base after a failed grasping attempt.
To enable these recovery behaviors, the robot must store and analyze prior failed executions when making future predictions.
To address this issue, \acro{} maintains \textbf{short-term memory} (Fig.~\ref{fig:method}, blue). The short-term memory stores the scene image, subtask, skill name, parameter, and the system-detected execution result (success/failure) at each prediction step for the \textit{current execution trial}.
This allows the VLM to reason over the entire execution history when predicting its next skill, crucial for long-horizon building-wide tasks.

The VLM may make reasoning errors, such as failing to consider object collisions with the environment, which could, for example, result in pushing an obstacle into a wall.
To allow~\acro{} to learn from these mistakes and reduce them~\cite{wang2024learning}, \acro{} maintains a \textbf{long-term memory} of previously collected prediction failures.
Erroneous VLM predictions, flagged by a human operator, are stored together with the decision context (user instruction, scene image, predicted subtask, skill name, and predicted parameters) and a VLM text analysis of the failure reason, serving as lessons for improving future predictions and reducing the dependency on human annotations.
\subsection{VLM-based Decision Making}
\begin{figure}[t!]
    \centering
    \includegraphics[trim={10px 0 660px 0},clip,width=\columnwidth]{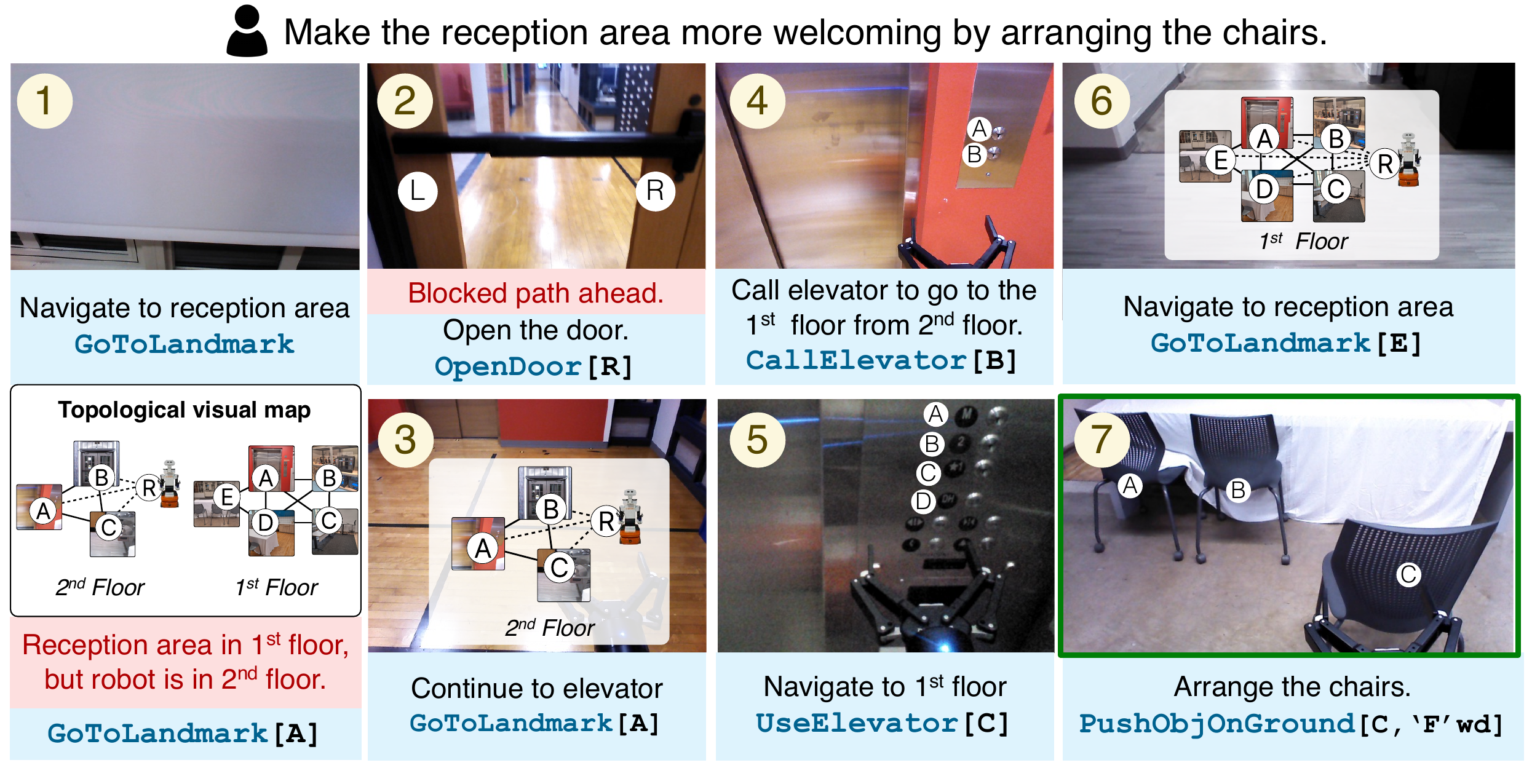}
    \vspace{-15px}
    \caption{\textbf{Execution trace for a user instruction.} For each decision step in the execution trace, we show the image observation with grounded skill parameters as markers, the skill name (blue text) and parameter executed, and a brief language description of the step (black) to improve readability. In trying to execute some skills, the robot fails (red text), in which case the VLM adaptively predicts the next skill.}
    \label{fig:exec_traces}
\end{figure}
\acro{} factorizes each decision-making step into two sequential steps (Fig.~\ref{fig:method}, yellow box). First, \acro{} exploits a VLM to perform perceptual and memory-based reasoning for predicting the next \textbf{subtask} to solve (e.g., ``go to a place where you can find a soda'') and for selecting the \textbf{skill} to handle it (e.g., \texttt{GoToLandmark}). Then, \acro{} queries the VLM to estimate the best \textbf{skill parameters} (e.g., \texttt{goal image of kitchen}) and instantiates the corresponding motor skill execution.

\paragraph{Subtask prediction and skill selection}
First, \acro{} needs to infer the next step of the task and what skill to use to solve it.
To that end, it integrates information about i) the language task instruction, ii) the description of robot skills, iii) the current scene (RGB) and robot state (location in the building), and iv) execution history (short-term memory) and failures from prior trials (long-term memory) into a \textit{unified multimodal VLM query}.

The VLM must also perceive the object distance information to infer the object's reachability, which is necessary for manipulation.
Naively prompting the VLM with depth images or point clouds does not yield desirable results because VLMs are not trained on those modalities. To allow the VLM to make informed decisions, \acro{} implements a multimodal Set-of-Mark (SoM)~\cite{som} prompting: foreground objects are marked with IDs in the input RGB image, and their distances to the robot, calculated using the perception system (Sec.~\ref{method:vision_mem}), are passed in the prompt.

\acro{} employs the resulting multimodal prompt to query the VLM-backbone about the next step in the task and the best-suited skill to solve it, requesting intermediate reasoning with Chain-of-Thought (CoT)~\cite{cot} for better accuracy.
With this VLM-backed procedure,~\acro{} can reason about long-horizon tasks, eliminating the need for a classical planner's computationally heavy search. 

\paragraph{Skill parameter estimation}
After predicting the subtask and skill to execute,~\acro{} predicts the parameters needed to correctly instantiate the skill and generate a sequence of low-level robot commands.
These parameters must be contextualized and grounded in the current scene for informed decision-making.
To that end, \acro{} creates a multimodal VLM query that includes i) the predicted subtask, ii) memory of previously made mistakes in predicting the skill parameters, and iii) an SoM-based visual grounding of the possible skill parameters with explanations of each marker (example in Fig.~\ref{fig:exec_traces}). For example, for the \texttt{MoveBase} skill, the possible parameters (forward, left, and right) 
are marked in the RGB image as arrow endpoints. 
For object-centric skills, the candidate parameters are generated by querying the GSAM with a skill-specific prompt (e.g., find `Buttons' for \texttt{UseElevator}).
\acro{} queries the VLM to choose the most promising skill parameter using CoT prompting for better reasoning about the subtask, scene, and past experiences. 
The execution outcome is stored in the short-term memory for iterative prediction.

%% file: 04_results.tex
\begin{table}[t!]
\caption{Success rate (\%) from $10$ trials in building-wide tasks}
\begin{adjustbox}{width=\columnwidth,center}
\centering
\small
\begin{tabular}{lcccccccc}
\toprule
& \multicolumn{3}{c}{Retrieve} & \multicolumn{3}{c}{Retrieve} & \multicolumn{1}{c}{Rearrange} & \\
& \multicolumn{3}{c}{marker} & \multicolumn{3}{c}{soda can} & \multicolumn{1}{c}{chairs} & Avg.\\
\cmidrule(l{4pt}r{4pt}){2-4} \cmidrule(l{4pt}r{4pt}){5-7} \cmidrule(l{4pt}r{4pt}){8-8} 
Building & \textit{B1} & \textit{B2} & \textit{B3} & \textit{B1} & \textit{B2} & \textit{B3} &  \textit{B1} & {(\%)}\\
\midrule
IM~\cite{huang2022inner} & $10$ & -- & -- & $0$ & -- & -- & $10$ & $6.7$ \\
COME~\cite{zhi2024closed} & $40$ & $30$ & $40$ & $\textbf{40}$ & $30$ & $30$ & $40$ & $35.7$ \\
\textbf{\acro{}} & $\textbf{60}$ & $\textbf{40}$ & $\textbf{50}$ & $\textbf{40}$ & $\textbf{50}$ & $\textbf{40}$ & $\textbf{50}$ & $\mathbf{47.1}$ \\
\bottomrule
\end{tabular}
\end{adjustbox}
\label{tab:main}
\end{table}

\section{Experimental Evaluation}
\label{s:exp}
To systematically evaluate the performance of~\acro{}, we benchmark \textbf{three} very long-horizon tasks up to $12$ skills long: \textit{Retrieving marker}, \textit{Retrieving diet soda can} and \textit{Rearranging chairs}. To test the system's language understanding capabilities, the tasks are specified using free-form text instructions, and the evaluations are averaged over three different task specifications for each task. For instance, in the retrieving marker task, we use ``\textit{I want to color the sky in my drawing. Can you get me a marker?}'', ``\textit{I want to color grass in my drawing. Can you get me a marker?}'' and ``\textit{I need to color some hearts. Can you get a marker for that?}'' (Refer to App.~\ref{app:evals} for details).

We test \acro{}'s building-wide MoMa capabilities by initializing the agent on different floors (possibly requiring elevator use), with randomized obstacles along the robot's path, such as closed doors, chairs, wet floor signs, and cardboard boxes. \acro{} must reason geometrically (e.g., selecting the correct arm to open a door or direction to push an obstacle) and semantically (e.g., avoiding wet floor signs).

We evaluate open-world generalization with multiple target objects per task---different brands of diet soda and markers, in random positions under varying degrees of clutter ($5$-$25$ distractor objects).
We test building-level generalization across three university buildings of different layouts, visual appearance, and room structures. Results are measured using $10$ trials per building per task (totaling $90+$ hours of evaluations). Since \acro{} demonstrated consistent performance on the first $2$ tasks across the $3$ buildings, we only evaluated our third task, rearranging chairs, in one building.

We compare~\acro{} with two baselines: \textbf{Inner-Monologue (IM)}~\cite{huang2022inner} that reasons using only a language scene description without RGB images or long-term memory, and \textbf{COME}~\cite{zhi2024closed}, which reasons using images, like \acro{}, but without long-term memory. 
Note that both methods were originally unsuitable for building-wide tasks due to the limited skill library and lack of memory, so we extended both with \acro{}'s diverse MoMa library and short-term memory, for fair comparison.
We instantiate \acro{} with GPT-4o for our experiments since we found it empirically to perform best.
We gather long-term memory experiences before evaluations and do not update them during evaluations.

Finally, in our evaluation, we complement the full-fledged long-horizon building-wide MoMa experiments described above by only assessing the parameter estimation for given skills in predefined scenes. To that end, we manually collected an offline dataset (\texttt{OfflineSkillDataset}) of around 120 images for three different skills and annotated them with ground truth parameters.

\begin{figure}[t!]
    \centering
    \includegraphics[width=\columnwidth]{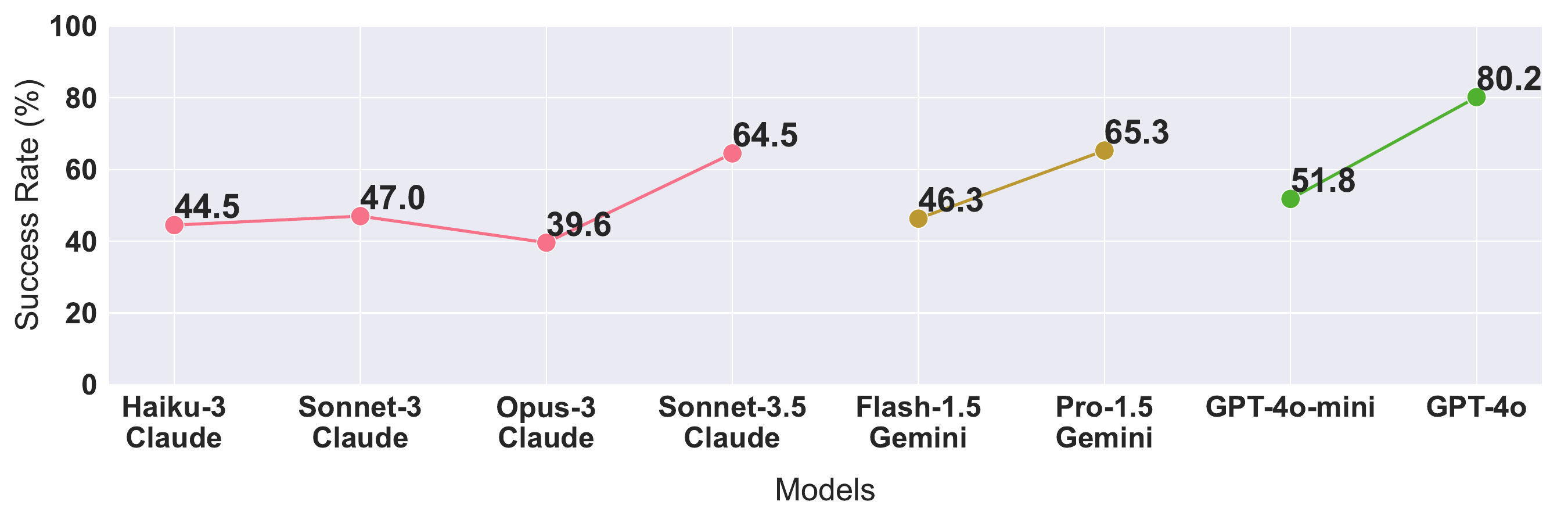}
    \vspace{-15px}
    \caption{Success rate (\%) of VLMs in predicting skill parameters. The models are arranged with increasing capabilities, measured as per the Vision Arena~\cite{chiang2024chatbot}, from left to right in \textit{each model series}.}
    \label{fig:diff_models}
\end{figure}

In our evaluation, we investigate the following questions:

\textbf{1) How does~\acro{} perform compared to other methods when addressing building-wide tasks?}
Table~\ref{tab:main} summarizes the results of our experiments. 
Inner-Monologue (IM) performs poorly compared to COME and \acro{}, indicating that multimodal prompting with visual information is critical to making the right decisions. Due to its poor performance, we restricted the evaluation of IM to that first building.
\acro{} scores best in all tasks and buildings and outperforms COME by $12.1\%$ on average, thanks to learning from past mistakes.
To directly compare the reasoning capability of these approaches over a diverse but fixed set of scenarios, we test each method against the ground truth skill parameters in \texttt{OfflineSkillDataset}.
\acro{} achieves $80.2\%$ success, outperforming COME ($72.6\%$) and Inner-Monologue ($61.7\%$), suggesting that \acro{}'s improvements in single-step skill parameter prediction help drive more successful executions in long-horizon tasks.

\textbf{2) How well does~\acro{} scale with increasing VLM capabilities?} In \acro{}, we sought to unify reasoning with a VLM-backbone to leverage advances in this fast-paced field. For that, \acro{} would have to work well with different VLMs and demonstrate improving capabilities with more powerful versions. Fig.~\ref{fig:diff_models} includes the results of our evaluation of \acro{} using Claude (Red), Gemini (Yellow), and GPT-4o (Green) model series as VLM-backbone on the \texttt{OfflineSkillDataset}. \acro{} works well with any VLM, and its performance with each model scales with VLM capabilities, highlighting the potential of \acro{} to leverage next-generation foundation models.

\begin{figure}[t!]
    \centering
    \includegraphics[width=\columnwidth, trim=100px 0 50px 0]{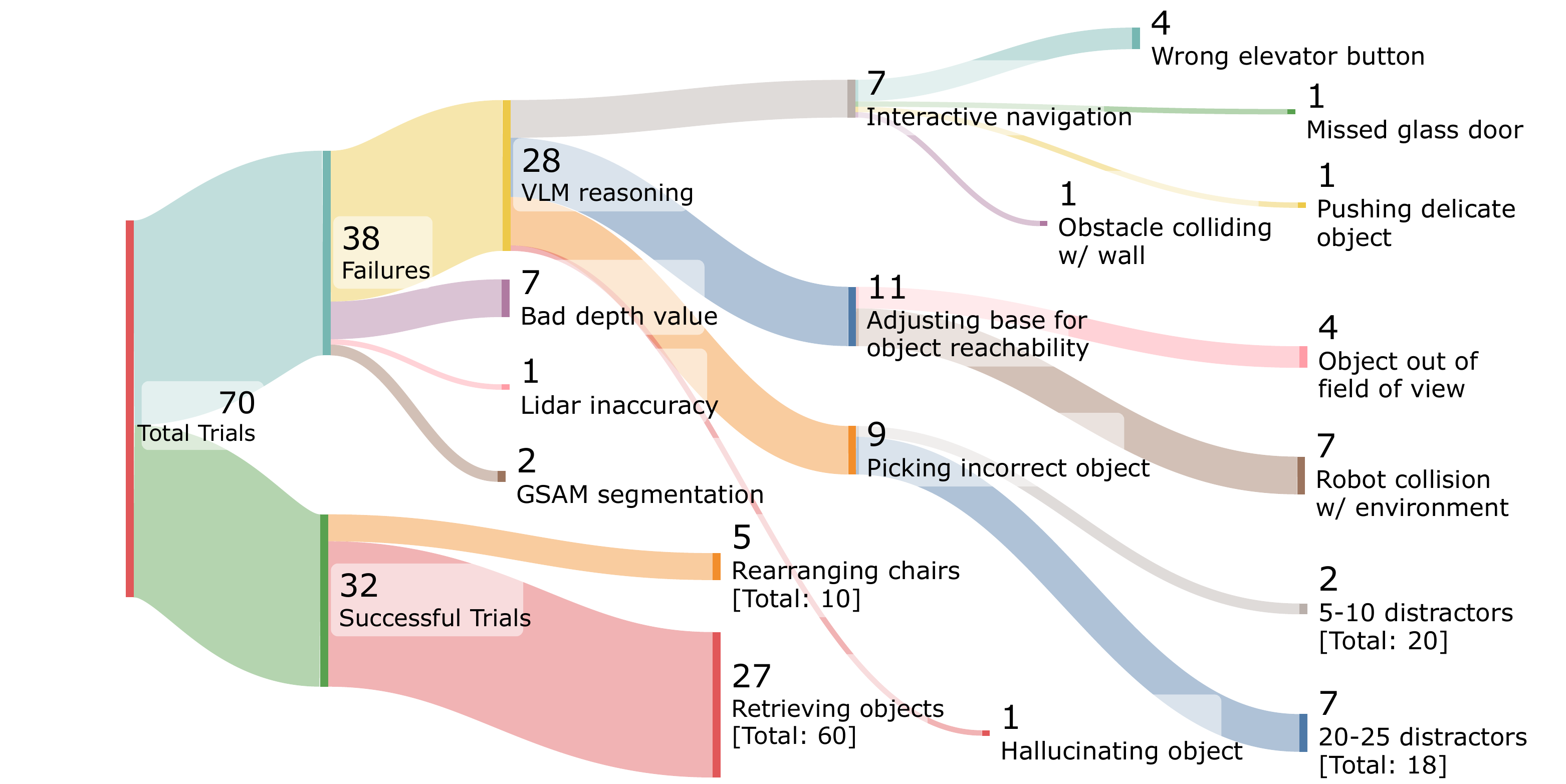}
    \caption{\textbf{Success and Failure Categorization} from 70 trials. Most failures involve VLM reasoning, such as predicting actions that lead to a collision with the environment, picking up the wrong object, especially when there are 20-25 distractors, or pressing the wrong elevator button.}
    \label{fig:failure}
\end{figure}

\textbf{3) What are the common failure modes in~\acro{}?}
We analyze and break down the failure cases observed during the evaluation of \acro{} in the three long-horizon tasks and present the results in Fig.~\ref{fig:failure}. 
Several of our errors ($10/38$) are caused by sensor failures (bad depth values, lidar failures) or GSAM segmentation mistakes, but most of them can be attributed to VLM reasoning mistakes, e.g., picking the wrong object, wrongly adjusting the base or failing to solve an interactive navigation problem~\cite{li2020hierarchical}.
We notice that these errors occur more often when \acro{} queries the VLM to localize markers demanding precise spatial understanding, such as elevator buttons, which often leads to pushing an incorrect one.
The VLM is also likelier to pick an incorrect object when there are many distractors ($20$-$25$, $38.9$\%) than a few ($5$-$10$, $10.0$\%), indicating limitations when reasoning with clutter, possibly due to presence of numerous markers. 

During evaluations, we noticed a surprising emergent behavior: \acro{} overcomes sensor failures using its diverse skill library. 
We observed that \acro{} learns to use \texttt{MoveBase} to a) reposition towards the elevator buttons, and b) approach an obstacle (chair), both after failed manipulation attempts due to NaN depth sensor values.

\textbf{4) How well does \acro{} align to human judgement?}
To analyze how well \acro{} decisions align with human judgment, we test on two completely unseen tasks without any prompt changes. These two tasks involve navigating to a completely new (shower) room: a) ``\textit{I spilled water on the floor. Can you pick up something to clean?}'', b) ``\textit{I dropped my phone in water and dried it off with a napkin, but it's still not working. Can you pick up something to get rid of all the moisture?}''. We then collect images of the execution using \acro{} and COME and annotate them with the task description and the parameterized skill selected by each one.
We ask $10$ participants to annotate each trajectory by selecting the description that best fits the result of the robot's last decision. Participants choose from five categories: \textbf{1}: Irrecoverable failure causing permanent damage to the robot or environment, \textbf{2}: Recoverable failure that can be corrected without lasting damage, \textbf{3}: Ill-specified decision, \textbf{4}: Sub-optimal task completion, \textbf{5}: Task completion that fully satisfies the user request. Participants respond on an online form with questions shuffled with a mix of decisions made by COME and \acro{}. 

\begin{figure}[t!]
    \centering
    \includegraphics[width=\columnwidth]{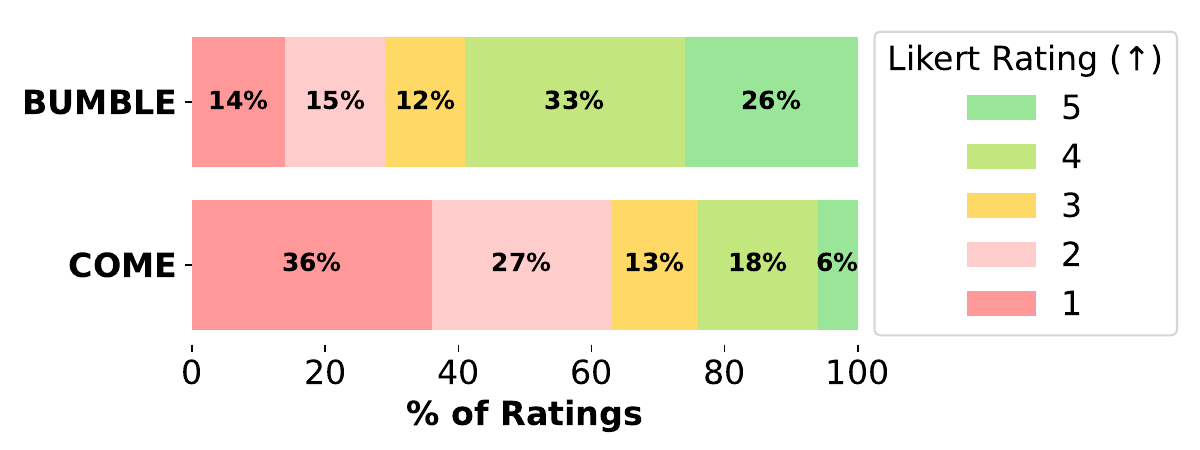}
    \vspace{-5px}
    \caption{\textbf{Human ratings on task completion and user satisfaction} ($n = 10$ participants, $10$ robot trials evaluated per participant per method) on a $5$-point Likert scale. \acro{} is rated better than COME at aligning to human preferences.}
    \label{fig:likert_plot}
\end{figure}

Fig.~\ref{fig:likert_plot} summarizes the results of our survey.
\acro{} achieves a Likert rating of $3.7$ out of $5.0$. Compared to COME (Avg. rating: $2.6$), \acro{} has $22\%$ points less irrecoverable and $12\%$ points less recoverable failures. 
Participants rate \acro{} as suboptimally completing the task $33\%$ of the time, such as successfully picking up an object, but not the one most suitable on the scene. We hypothesize this is due to greedy plans generated by VLMs, which often pick up the nearest, most visible object. 
Incorporating multi-step reasoning using VLMs to overcome greedy behavior is an exciting future direction for~\acro{}. 

\textbf{5) What is the contribution of CoT and SoM to \acro{}?}
We ablate two components of the \acro{} VLM interface: Chain-of-Thought reasoning and Set-of-Mark prompting to measure their effect on performance.
Not using CoT decreases performance by $19.5\%$ points, down to $60.7\%$, underscoring the importance of having intermediate reasoning steps for decisions in building-wide tasks.
Removing SoM and instead using the VLM to predict a desired object description in natural language, then asking GSAM to segment this object~\cite{stone2023open}, decreases performance by $31\%$ points, highlighting GSAM's difficulty in segmenting specific object instances such as ``\textit{Diet} Dr. Pepper'' when a non-diet Dr. Pepper can is present (See Table~\ref{tab:skill_param}).

%% file: 05_conclusion.tex
\section{Conclusion}
We demonstrate that a VLM-backed policy, acting as a central reasoning module equipped with a diverse skill library, memory, and open-world perception system, can effectively solve building-wide mobile manipulation tasks. Comprehensive ablations and experiments demonstrate the potential of \acro{} as a stepping stone for achieving autonomous service robots in buildings.

\textbf{Limitations and Future Work}:
\acro{}'s long-term memory storing previous failures can become computationally intractable with time; using a dynamic retrieval process from a large storage space or compressing these experiences in learnable weights can be explored in future work.
Furthermore, \acro{} assumes access to landmarks with goal images in the \texttt{GoToLandmark} skill, which requires human effort in collecting the landmarks. Autonomous exploration with landmark frame selection can be explored to offload human effort.
Lastly, while~\acro{} can be easily interfaced with semantically meaningful skills, adding skills or parameters that cannot be easily expressed via language~\cite{sharma2019dynamics,pertsch2021guided} requires additional machinery for video conditioning or specialized weights to allow such interfacing.
\section{Acknowledgments}
We thank Shivin Dass for developing the robot system infrastructure. We also thank Ben Abbatematteo, Shuijing Liu, and Mingyo Seo for their feedback on the manuscript.
We thank all UT Austin Robot Perception and Learning Lab and Robot Interactive Intelligence Lab members for their invaluable feedback on~\acro{}. 
This work was partially supported by the National Science Foundation (FRR2145283, EFRI-2318065), the Office of Naval Research (N00014-22-1-2204, N00014-24-1-2550), and the DARPA TIAMAT program (HR0011-24-9-0428).

%% file: appendix.tex
\section{Appendix}
\begin{table*}[ht!]
\centering
\small
\begin{tabular}{l|c|c|c|c|c}
\toprule
Skill & \acro{} & COME & Inner & \acro{} & \acro{} \\
Parameter & & & Monologue & w/o CoT \cite{cot} & w/o SoM \cite{som}\\ 
\midrule
\texttt{Pickup [Obj.]} (5-10 distractors) & $80.0$ & $80.0$ & $65.0$ & $80.0$ & $50.0$ \\
\texttt{Pickup [Obj.]} (20-25 distractors) & $65.0$ & $65.0$ & $65.0$ & $60.0$ & $40.0$ \\
\texttt{PushObjectOnGround [Obj., Dir.]} & $81.0$ & $70.4$ & $56.8$ & $64.3$ & $81.8$ \\
\texttt{CallElevator [Button Location]} & $95.0$ & $75.0$ & $60.0$ & $40.0$ & $25.0$ \\
\midrule
Average & $\mathbf{80.2}$ & $72.6$ & $61.7$ & $60.7$ & $49.2$ \\
\bottomrule
\end{tabular}
\caption{Success Rate (\%) in predicting skill parameters using \texttt{OfflineSkillDataset}.}
\label{tab:skill_param}
\end{table*}
\subsection{Method details}\label{app:method}
\paragraph{\textbf{Perception System}}
We use a strong open-world segmentation model, GSAM, to extract interactable objects from the scene image with pixel-level accuracy. GroundingDINO~\cite{liu2023grounding} (\verb|groundingdino_swinb_cogcoor|) provides the bounding box and SAM-HQ~\cite{sam_hq} (\verb|sam_hq_vit_b|) serves as the segmentation model. We borrow the code from ORION~\cite{zhu2024orion}.
\paragraph{\textbf{Skill Library}}
\acro{} has a diverse skill library to operate effectively in buildings.
For each skill, we provide the description generating the parameters, type of parameters, name \& description of the skill provided to the VLM for skill selection, and description of parameter mapping to low-level robotic actions (See Fig.~\ref{fig:exec_traces} for prompt images). \\
\texttt{\textbf{GoToLandmark}} skill enables the robot to navigate between locations on the same floor using a topological visual map with landmark images as nodes and a 2D occupancy map to generate trajectories. The VLM receives all landmark images with markers and the corresponding floor number. Based on the image semantics, it selects the most suitable landmark for the subtask (including the elevator landmark, if necessary).
\\
\textit{Skill parameter(s)}: Landmark image
\\
\textit{Skill description prompt for skill selection}:
\begin{lstlisting}
skill_name: goto_landmark
arguments: Selected landmark image from the environment from various options.
description: Navigates to the landmark in the environment. For instance, bedroom, kitchen, tool shop, etc.
\end{lstlisting}
\textit{Parameter to low-level robot commands}:
The landmark images have a pre-defined mapping to the corresponding 2D pose on the occupancy map, which acts as the target pose. We use a global and local planner to generate a trajectory from the current robot pose to the target pose using the 2D occupancy map and the robot's front 2D lidar.
\\
\texttt{\textbf{NavigateNearObj}} skill enables the robot to traverse near an object that is visible to the robot. We use a GSAM query \verb|all objects| to segment objects in the scene. The segmentations are overlayed with markers on the image.\\
\textit{Skill parameter(s)}: Object segmentation
\\
\textit{Skill description prompt for skill selection}:
\begin{lstlisting}
skill_name: navigate_to_point_on_ground
arguments: object
description: Moves the robot to a point near the selected object. This skill can be used to move to a point in the room to perform a task, example, navigating near the toaster to make a toast.
\end{lstlisting}
\textit{Parameter to low-level robot commands}:
Object segmentation is used to calculate the object's pose relative to the robot. The nearest point on the ground to the object is converted to a 2D pose on the map, which in turn is used to generate low-level robot base commands.
\\
\texttt{\textbf{MoveBase}} skill equips the robot to adjust the base by $30$ cm in one of the four directions: forward, left, right, and backward. The pose of the robot base after moving $30$ cm in each direction is calculated and overlayed on the scene image.
\\
\textit{Skill parameter(s)}: Direction
\\
\textit{Skill description prompt for skill selection}:
\begin{lstlisting}
skill_name: move_base
arguments: direction
description: Moves the robot base either forward, backward, left, or right by 0.3 meters, w.r.t. the camera view. This skill should only be used to adjust the base of the robot in the final few meters of navigation and not to traverse between rooms.
\end{lstlisting}
\textit{Parameter to low-level robot commands}:
The robot's pose relative to the map frame after moving $30$ cm in the selected direction is calculated and used as the target base pose.
\\
\texttt{\textbf{Pickup}} skill allows the robot to pick up an object using the left arm. We use a GSAM query \verb|all objects| to segment objects in the scene. The segmentations are overlayed with markers on the image.
\\
\textit{Skill parameter(s)}: Object segmentation
\\
\textit{Skill description prompt for skill selection}:
\begin{lstlisting}
skill_name: pick_up_object
arguments: object_of_interest
description: pick_up_object skill moves its arms to pick up object_of_interest. The pick_up_object skill can only pick up objects within arm reach and does not control the robot base. The robot cannot pick up heavy objects like chairs, tables, etc.
\end{lstlisting}
\textit{Parameter to low-level robot commands}:
The object pose is calculated relative to the robot using the segmentation mask, which then helps determine an approach and a goal pose for the left-arm gripper. Inverse Kinematics is used to find the corresponding joint configuration, which is then achieved via the joint controller.
\\
\texttt{\textbf{PushObjOnGround}} skill enables the robot to push a visible object forward, left, or right. We query the VLM twice: a) to select the object to push where the possible options are generated using a GSAM query (\verb|all objects|), and b) to determine the push direction, object position after pushing is approximated and overlayed on the image.
\\
\textit{Skill parameter(s)}: Object segmentation, and direction
\\
\textit{Skill description prompt for skill selection}:
\begin{lstlisting}
skill_name: push_object_on_ground
arguments: object, direction
description: Pushes an object on the ground one of the directions. The robot can push objects that are 2-3m away from the robot. The skill decides which object and direction to push. Example use cases: pushing obstacle to clear the pathway for the robot to go forward, pushing objects for rearrangement, etc.
\end{lstlisting}
\textit{Parameter to low-level robot commands}:
The selected object's pose is used to determine the target 2D pose for the robot base via heuristics. The robot then moves to the target pose and executes a predefined arm trajectory based on the chosen direction.
\\
\texttt{\textbf{OpenDoor}} enables the robot to open push-doors using the left or right arm. Two markers indicating left or right are overlayed on the image (central left, central right image positions).
\\
\textit{Skill parameter(s)}: Left or Right
\\
\textit{Skill description prompt for skill selection}:
\begin{lstlisting}
skill_name: open_door
arguments: door_side
description: Opens the door by pushing on a certain side of the door (left or right). The robot moves forward with its door_side arm out to push open the door.
\end{lstlisting}
\textit{Parameter to low-level robot commands}: The robot aligns itself with the robot door. It moves the selected arm to a predefined joint configuration and then moves toward the door to push it open using the forearm of the selected arm as the contact point.
\\
\texttt{\textbf{CallElevator}} allows the robot to call the elevator to the current floor and move inside the elevator. The buttons are segmented using GSAM query \verb|buttons|.
\\
\textit{Skill parameter(s)}: Button Segmentation
\\
\textit{Skill description prompt for skill selection}:
\begin{lstlisting}
skill_name: call_elevator
arguments: button position depending whether you want to go to a floor above or below.
description: Equips the robot with calling the elevator capability. The robot will push the button selected in the argument to call the elevator in the current floor. The subtask must indicate the current floor number and the destination floor number to go to. Example: 'Go to the second floor from first floor.'
\end{lstlisting}
\textit{Parameter to low-level robot commands}: The button pose is calculated using the selected segmentation mask. The button is pressed until reaching a force threshold, measured by the force-torque sensor on the robot's right wrist. After pressing the button, the robot moves inside the elevator to a predefined map pose.
\\
\texttt{\textbf{UseElevator}} allows the robot to use the elevator to go to a target floor and then move outside the elevator. The buttons are
segmented using GSAM query buttons.
\\
\textit{Skill parameter(s)}: Button Segmentation
\\
\textit{Skill description prompt for skill selection}:
\begin{lstlisting}
skill_name: use_elevator
arguments: button position of the floor to go to.
description: Equips the robot with using elevator capabilities. The robot will push the button selected in the argument. This skill is used to change the floor of the robot after calling the elevator. The subtask must indicate the desired floor number to go to.
\end{lstlisting}
\textit{Parameter to low-level robot commands}: The button pose is calculated using the selected segmentation mask. The button is pressed until a force threshold is crossed, measured by the force-torque sensor on the robot's right wrist. After pressing the button, the robot moves outside the elevator and localizes itself on the new floor 2D map.
\paragraph{\textbf{ROS Packages}}
For these skills, we use the following ROS packages, configured for the Tiago robot: \verb|gmapping| (2D occupancy map generation), \verb|amcl| (Robot localization), \verb|move_base| (to reach the base pose relative to the 2D map). The \verb|move_base| package uses \verb|global_planning| ($A^{*}$ global path planner), \verb|teb_local_planner| (Time Elastic Band local planner). Finally, we use the joint controller configured for Tiago: \url{wiki.ros.org/Robots/TIAGo/Tutorials/trajectory_controller}. We borrow the codebase from TeleMoMa~\cite{dass2024telemoma}.
\paragraph{\textbf{Memory}}
The short-term memory enables the robot to track state-action history and overcome past failures. These failures, detected by the skills, include the depth sensor returning NaN values for a selected object, failure to find a valid inverse kinematics solution for arm pose, and inability to find a collision-free path due to obstacles like a chair, wet floor signs, or closed doors.

The long-term memory allows the VLM to learn from its past mistakes. These experiences are collected offline and are not updated during evaluations. We collect small examples of $5$ instances for each skill and annotate them with ground truth answers. We compare the predicted and ground truth answers, and only the erroneous predictions are retained. This yields $1$-$3$ examples for the \texttt{MoveBase}, \texttt{UseElevator}, \texttt{PushObjOnGround} skill parameters, and for the skill selection.

\subsection{Experimental Evaluations}\label{app:evals}
\paragraph{Task Specifications} For each task, we evaluate the three task specifications highlighted below to demonstrate the robustness of our system to diverse user language instructions.\\
\textbf{Retrieve diet soda can}:
1) ``Could you grab me a drink that is low in calories?''
2) ``Any chance you can find me a sugar-free soda?''
3) ``I want something fizzy to drink, but I am on diet. Can you help me with that?''
\\
\textbf{Retrieve colored marker}:
1) ``I want to color the sky in my drawing. Can you get me a marker?''
2) ``I want to color grass in my drawing. Can you get me a marker?''
3) ``I need to color some hearts. Can you get a marker for that?''
\\
\textbf{Rearrange chairs}:
1) ``Could you make the seating chairs in the reception area more orderly?''
2) ``Make the reception area more welcoming by arranging the chairs''
3) ``Can you help me arrange the chairs in the reception area?''
\paragraph{VLM checkpoints} We demonstrate in Fig.~\ref{fig:diff_models} that \acro{} is a general framework that leverages the rapidly advancing capabilities of VLM. We provide the VLM checkpoint details for reproducibility: GPT4o (\textit{gpt-4o-2024-05-13}), GPT4o-mini (\textit{gpt-4o-mini-2024-07-18}), Pro-1.5-Gemini (\textit{Gemini-1.5-Pro-Exp-0827}), Flash-1.5-Gemini (\textit{Gemini-1.5-Flash-Exp-0827}), Haiku-3-Claude (\textit{claude-3-haiku-20240307}), Sonnet-3-Claude (\textit{claude-3-sonnet-20240229}), Opus-3-Claude (\textit{claude-3-opus-20240229}), Sonnet-3.5-Claude (\textit{claude-3-5-sonnet-20240620}).